\documentclass[conference]{IEEEtran}
\IEEEoverridecommandlockouts
\usepackage[T1]{fontenc}
\usepackage{cite}
\usepackage{amsmath,amssymb,amsfonts}
\usepackage{algorithmic}
\usepackage[left=0.66in, right=0.66in, top=0.74in, bottom=0.94in]{geometry}
\usepackage{graphicx}
\usepackage{textcomp}
\usepackage{xcolor}
\usepackage{adjustbox}
\usepackage{multirow}
\usepackage{booktabs}
\usepackage{caption} 
\usepackage{array} 
\usepackage{tabularx} 
\usepackage{stmaryrd} 
\usepackage{multicol} 
\usepackage[headerAttributes,texMathDollars]{markdown}
\markdownSetup{
    renderers={
	heading* = {{\bf #1}}
    }
}
\usepackage{textcomp}
\usepackage{scalerel}
\def\memo{\scalerel*{\includegraphics{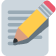}}{\textrm{\textbigcircle}}}
\def\speak{\scalerel*{\includegraphics{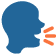}}{\textrm{\textbigcircle}}}
\def\robot{\scalerel*{\includegraphics{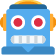}}{\textrm{\textbigcircle}}}
\usepackage{tcolorbox}
\usepackage[frozencache,cachedir=.]{minted}
\tcbuselibrary{minted}
\tcbuselibrary{skins}
\tcbuselibrary{breakable}
\def\BibTeX{{\rm B\kern-.05em{\sc i\kern-.025em b}\kern-.08em
    T\kern-.1667em\lower.7ex\hbox{E}\kern-.125emX}}
\definecolor{openai-primary}{HTML}{0FA37F}
\definecolor{openai-gray}{HTML}{E6F3F3}
\definecolor{user-request-primary}{HTML}{F4AA40}
\definecolor{user-request-gray}{HTML}{FDF3E4}
\definecolor{json-primary}{HTML}{6A33CD}
\definecolor{json-gray}{HTML}{F6F2FC}
\colorlet{gray-background}{black!5!white}
\NewTotalTColorBox{\URBox}{m}{enhanced,
	breakable,
	drop fuzzy shadow,
	colback=gray-background,
	colframe=black,
	boxsep=.0pt,
	left=3mm,
	fontlower=\sffamily\scriptsize,
	halign=justify,
	sidebyside,
	sidebyside gap=1.5mm,
	sidebyside align=top,
	lower separated=false,
	lefthand width = 5mm,
	leftrule=3mm}{\speak \tcblower #1}
\NewTotalTColorBox{\Request}{m}{
	boxrule=0pt,
	boxsep=.0pt,
	frame empty,
	left=3mm,
	right=3mm,
	fontlower=\sffamily\scriptsize,
	halign=justify,
	sidebyside,
	sidebyside gap=1.5mm,
	sidebyside align=top,
	lower separated=false,
	lefthand width = 5mm,
	bottom=0pt
}{\speak \tcblower \textbf{Request: }#1}
\NewTotalTColorBox{\Answer}{m}{
	boxrule=0pt,
	boxsep=.0pt,
	frame empty,
	left=3mm,
	right=3mm,
	fontlower=\sffamily\scriptsize,
	halign=justify,
	sidebyside,
	sidebyside gap=1.5mm,
	sidebyside align=top,
	lower separated=false,
	lefthand width = 5mm,
	top=0pt
}{\robot \tcblower \textbf{Answer: }#1}
\NewTotalTColorBox{\ConvBox}{m m}{enhanced,
	boxsep=0pt,
	breakable,
	drop fuzzy shadow,
	colback=gray-background,
	colframe=black,
	leftrule=3mm,
	left=0pt,
	top=0pt,
	bottom=0pt,
	right=0pt,
}{\Request{#1}\Answer{#2}}
\newtcolorbox{markdownbox}[2][]{colback=gray!10, 
  colframe=black,
  title=#2, #1, enhanced,
  sharp corners,
  fonttitle=\bfseries, attach boxed title to top center={yshift=-2mm}
}
\NewTotalTColorBox{\InstructBox}{m}{%
	enhanced,
	breakable,
	drop fuzzy shadow,
	leftrule=3mm,
	boxsep=.0pt,
	left=3mm,
	right=3mm,
	halign=justify,
	fontlower=\sffamily\scriptsize,
	colback=gray-background,
	colframe=black,
	sidebyside,
	sidebyside gap=1.5mm,
	sidebyside align=top,
	lower separated=false,
	lefthand width = 5mm,
	}{\memo \tcblower{#1}}
\newtcolorbox{JsonBox}[2][]{%
  enhanced,
  breakable,
  colback=gray!10,
  colframe=black,
  fonttitle=\ttfamily\bfseries,
  title={#2},
  listing only,
  boxsep=0pt,
  left=5pt,
  right=5pt,
  top=5pt,
  bottom=5pt,
  #1
}

\newtcbinputlisting{\InstructBoxFile}[2][]{%
    listing engine=minted,
enhanced,
breakable,
drop fuzzy shadow,
listing file={#2},
leftrule=2.5mm,
boxsep=-2pt,
    left=5pt,
colback=gray-background,
colframe=black,
    comment side listing,
sidebyside gap=1.2mm,
sidebyside align=top,
lower separated=false,
lefthand width = 4mm,
minted language=text,
    minted options={fontsize=\scriptsize,breaklines,fontfamily=helvetica},
    comment=\memo,
#1}
\usepackage{tikz}
\usetikzlibrary{positioning,shapes.geometric,arrows.meta,shadows.blur,backgrounds,fit,calc}
\tikzset{
    basicblock/.style={
        rectangle,
        rounded corners=5,
        inner sep=6pt,
        text centered,
        minimum width=1.6cm,
        minimum height=.8cm,
        font = \sffamily\scriptsize,
        line width=1.5pt
    }
}

\tikzset{
	basicblockshadow/.style={
	rectangle,
	rounded corners=5,
	inner sep=6pt,
	blur shadow={
		shadow blur steps=5,
		shadow blur radius=1pt, 
		shadow xshift=3pt,
		shadow yshift=-3pt,
		shadow opacity=22
	},  
	text centered,
	minimum width=1.6cm,
	minimum height=.8cm,
	node font = \normalsize,
	line width=1.5pt
	}
}

\tikzset{
	myarrow/.style={
		-{Latex[round]},
		thick
	}
}

\usepackage{pifont}
\begin{document}

\title{Large Language Models for Power Scheduling:\\ A User-Centric Approach \\
\thanks{The authors acknowledge the KU-TII 6G Chair on Native AI.}
}

\author{
    \IEEEauthorblockN{Thomas Mongaillard\IEEEauthorrefmark{1}\IEEEauthorrefmark{2}, Samson Lasaulce\IEEEauthorrefmark{2}\IEEEauthorrefmark{1}, Othman Hicheur\IEEEauthorrefmark{2}\IEEEauthorrefmark{3}, Chao Zhang\IEEEauthorrefmark{2}\IEEEauthorrefmark{4}, Lina Bariah \IEEEauthorrefmark{5}\IEEEauthorrefmark{2}, \\Vineeth S. Varma\IEEEauthorrefmark{1}, Hang Zou \IEEEauthorrefmark{6}, Qiyang Zhao \IEEEauthorrefmark{6}, and Merouane Debbah\IEEEauthorrefmark{2}\IEEEauthorrefmark{6}}
    \IEEEauthorblockA{\IEEEauthorrefmark{1} Université de Lorraine, CNRS, CRAN, F-54000 Nancy, France.}
    \IEEEauthorblockA{\IEEEauthorrefmark{2} KU 6G Research Center, Khalifa University, Abu Dhabi, UAE.}
    \IEEEauthorblockA{\IEEEauthorrefmark{3} Ecole Polytechnique, Paris, France. \,\,\, \IEEEauthorrefmark{4}Central South University, Changsha, China.}
    \IEEEauthorblockA{\IEEEauthorrefmark{5}Open Innovation AI, Abu Dhabi, UAE. \,\,\, \IEEEauthorrefmark{6}TII, Abu Dhabi, UAE}
}

\maketitle

\begin{abstract}
While traditional optimization and scheduling schemes are designed to meet fixed, predefined system requirements, future systems are moving toward user-driven approaches and personalized services, aiming to achieve high quality-of-experience (QoE) and flexibility. This challenge is particularly pronounced in wireless and digitalized energy networks, where users' requirements have largely not been taken into consideration due to the lack of a common language between users and machines. The emergence of powerful large language models (LLMs) marks a radical departure from traditional system-centric methods into more advanced user-centric approaches by providing a natural communication interface between users and devices. In this paper, for the first time, we introduce a novel architecture for resource scheduling problems by constructing three LLM agents to convert an arbitrary user's voice request (VRQ) into a resource allocation vector. Specifically, we design an LLM intent recognition agent to translate the request into an optimization problem (OP), an LLM OP parameter identification agent, and an LLM OP solving agent. To evaluate system performance, we construct a database (\textcolor{blue}{\textsf{EVRQ}}) of typical VRQs in the context of electric vehicle (EV) charging. As a proof of concept, we primarily use Llama 3 8B. Through testing with different prompt engineering scenarios, the obtained results demonstrate the efficiency of the proposed architecture. The conducted performance analysis allows key insights to be extracted. For instance, having a larger set of candidate OPs to model the real-world problem might degrade the final performance because of a higher recognition/OP classification noise level. [Paper codes and video\footnote{\texttt{https://github.com/thomasmong/llm-power-scheduling}}].

\end{abstract}


\section{Introduction}
\label{sec:intro}

With the rapid evolution of complex systems across various domains, the need for sophisticated scheduling schemes to efficiently manage the system resources and meet certain requirements has become increasingly critical. While recently we have witnessed noticeable advancements in the development of scheduling algorithms (particularly with leveraging artificial intelligence (AI)-driven methods), it is essential to emphasize that these traditional algorithms are often designed to satisfy predefined constraints that are inherently tied to the specific system they serve. Although they have proven effective in many scenarios, they often fail to perform well when faced with the dynamic and personalized demands of modern users.

In addition to the fact that they might be uncapable of satisfying the level of quality-of-experience (QoE) imposed by services that are increasingly oriented towards satisfying user demands, conventional methods might result in high complexity and not necessarily lend themselves into an optimum solution with respect to energy-efficiency in personalized scenarios. Taking the example of energy management, a key aspect underpinning reliable and sustainable operations in many systems, including wireless communication networks, autonomous vehicular systems, smart grids, etc., humans remain largely out of the loop. For instance, for heating or air conditioning (AC) systems, humans typically just provide the target temperature, and a more or less advanced regulation algorithm does all the rest. This approach operates independently of individual user preferences, or potential changes in energy availability or costs. Consequently, it may not always optimize for energy-efficiency or user comfort, demonstrating a clear limitation.

 
The reason for this gap between humans and algorithms is twofold. First, most humans are not able to model mathematically a real-world problem and solve it, which is why algorithms are typically tasked with making most decisions. Second, humans cannot communicate easily or not at all with algorithms, machines, or programs. However, the emergence of advanced natural language processing (NLP) tools, such as large language models (LLMs), is completely changing the paradigm. In particular, LLMs enable intuitive and effective human-machine interactions, transforming the operation of complex infrastructures, such as energy and wireless networks, into more responsive and user-centric solutions. 

It should be noted that the problem of power scheduling is a key in both wireless networks and in digitalized energy networks. Many wireless resource management problems and home energy management problems can be formulated as power scheduling problems \cite{fattah-wireless-2002,elbatt-twc-2004,hohlt-isipsn-2004,radunovic-jsac-2004,makhamed-elsevier-2019}. The main goal of this paper is to leverage the capabilities of LLMs in order to enable the machine to convert a voice request (VRQ) from a human user into a power scheduling vector. For example, for the case study under consideration in this paper (namely electric vehicle -EV- charging), such a request can be: "Charge my EV for tomorrow at 6 a.m. while managing its battery lifetime". For a cell phone, it might be "Adapt your transmit power to minimize electromagnetic exposure while guaranteeing my SMS messages always go through". \textit{
The novel approach to power scheduling introduced and developed in this paper is to exploit the knowledge the LLM has acquired during pre-training and auxiliary instructions to both model and solve the problem at hand.} It is important to highlight that, although we develop this approach for the particular case of the power scheduling problem, such an approach can be generalized to help humans solve a wide range of real-world problems by leveraging mathematical modeling, reasoning, and solving capabilities.

Nowadays, there are several services, e.g., Amazon Alexa, Google Smart Home, and Siri-based Apple Smart Home, with advanced interfaces that allow humans to "talk" to devices such as electrical appliances. However, in all these solutions, the employed deep learning algorithms only act as mere classifiers, i.e., they classify the user’s request into a given control action (e.g., switch a given electric appliance on). Therefore, the implemented deep learning schemes do not try to model mathematically or interpret the physical problem at hand, and thus they do not attempt to solve it by exploiting the interpretation, reasoning, and planning capabilities that LLMs-at least partially-have (see, e.g., \cite{GPT-3}\cite{Llama2}). Additionally, existing deep learning solutions are trained for particular tasks, and hence, they do not generalize to a wider range of tasks, as would be encountered when treating VRQs made by a human user. Therefore, the standpoint of this paper is original in the sense that the proposed architecture aims at exploiting LLMs to mathematically model the physical problem at hand and solve it. To our knowledge, the closest literature to this approach is given by the literature of math word problems (see e.g., \cite{koncel-NAC-2016,wang-proc-2017,heyueya-2023,OPRO}). In this literature, the real-world problem is assumed to be perfectly specified by (textual) natural language. The proposed framework is novel in the following ways. Unlike the aforementioned literature, the current framework does not rely on the assumption that user requests are structured in a specific format; instead, they can be arbitrary. This framework is specifically tailored to address power scheduling tasks, which has not been tackled in the literature in the context of LLMs, yet. For the first time, the proposed approach exploits the capabilities of LLMs to perform both, modeling and solving power scheduling problems. The aim is not to provide a power scheduling scheme that outperforms existing ones in terms of predefined performance metric, but rather to revolutionize how optimization problems (OPs) are formulated and solved for the sake of generating the recommended power scheduling vector.

 To achieve our goal, we propose an LLM multi-agent architecture which allows a human VRQ to be converted to an OP whose solution is a power scheduling vector (Vec). The approach of multiple LLM agents has been developed recently (see e.g., \cite{talebirad-MA-2023,babyagi-2023,autogpt-2023,COE}) and consists in specializing an LLM for a given range of tasks (experts) and associating the LLM agents for a global task. 
The main contributions of this paper are summarized as follows: $\bullet$ We develop a new \textbf{methodology} (based on LLM) to design a scheduling scheme for power management systems, in which the system requirements are acquired from the user through VRQs (Sec.~\ref{sec:problem-statement}); $\bullet$ We propose a novel \textbf{architecture} (Sec.~\ref{sec:architecture}) comprising the design of three LLM agents, with the aim to perform user-driven power scheduling, i.e., an intent recognition agents (to identify the best formulation of the mathematical problem given the VRQ from the user), a parameter identification agent (to determine the required parameters for the OP from the VRQ as well as the physical system), and an OP solving agent (to solve the formulated problem through a bank of solving functions in which the LLM agent assists the solver in the initialization phase through particular prompts); $\bullet$ The proposed architecture is partially implemented mainly for Llama3 and evaluated through a thorough \textbf{performance analysis} (Sec.~\ref{sec:perf-evaluation}); $\bullet$ As effective accuracy measure, we construct a \textbf{database} of possible user VRQs related to EV charging (\textcolor{blue}{\textsf{EVRQ}}), with the corresponding request complexity, ground-truth OP/OCP classes, and optimal solutions. This database enables us to evaluate the performance of our framework by comparing the generated results against these benchmarks, using intent recognition accuracy (IRA) and final utility (Sec.~\ref{sec:perf-evaluation}).

\section{Problem Statement}
\label{sec:problem-statement}

\begin{figure*}
    \centering
 \includegraphics[width=0.8\linewidth, height=0.31\linewidth]{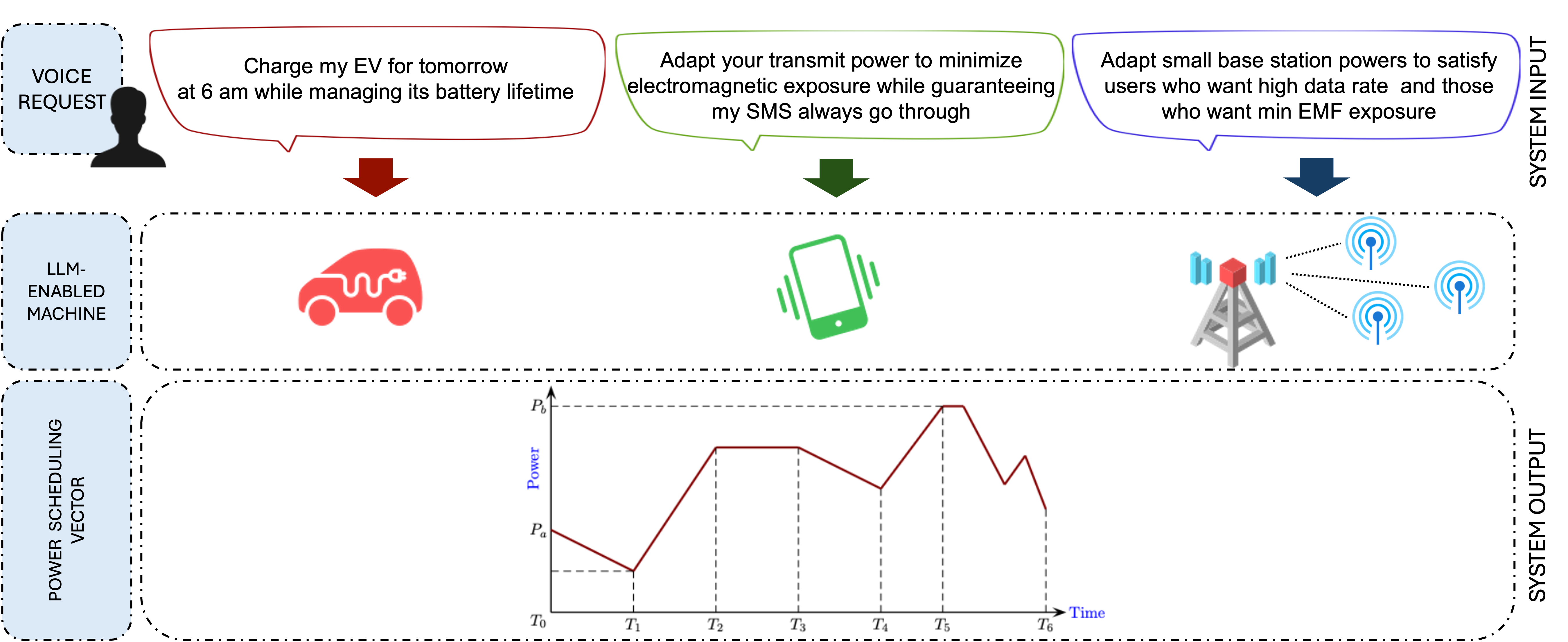}
    \caption{Use-Cases of the Proposed Intelligent Power Scheduling System}
    \label{fig:use-cases}
\end{figure*}

In this work, we develop an LLM-based converter which takes as input a voice/text request from the user (see examples of use cases in Fig. \ref{fig:use-cases}) and transforms it into a power consumption scheduling vector (Vec)
\begin{equation}
    x = (x_1,x_2, ...,x_T)
\end{equation}
with $\forall t \in \{1,...,T\},  0 \leq x_{\min} \leq x_t \leq x_{\max} $, $T$ being the number of time-slots over which the power is scheduled.


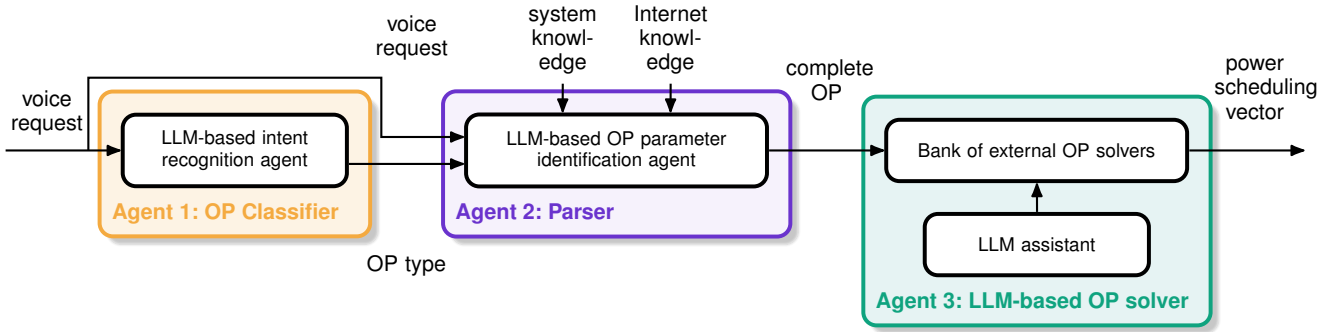
\begin{figure*}[ht!]
    \centering
    \begin{tikzpicture}[font=\small, scale=1.02, transform shape] 
    \def\pad{1.5}
    \node (OR) {};
    \node[basicblock,draw=black,fill=white, text width=2.5cm, right = \pad of OR] (classify) {LLM-based intent recognition agent};
    \node[basicblock,draw=black,fill=white, text width=3.5cm, right = \pad of classify] (parse) {LLM-based OP parameter identification agent};
    \node[basicblock,draw=black,fill=white, text width=3.5cm, right = \pad of parse] (solve) {Bank of external OP solvers};
    \node[basicblock,draw=black,fill=white, text width=2.5cm, below = \pad / 4 of solve] (assist) {LLM assistant};
    \node[right = \pad of solve] (END) {};
    \node[above = \pad/4 of parse, font=\footnotesize\sffamily, xshift=-20pt, text width=3em, align=center] (sysk) {system knowledge};
    \node[above = \pad/4 of parse, font=\footnotesize\sffamily, xshift=20pt, text width=3em, align=center] (webk) {Internet knowledge};

    \node[basicblock, below = \pad / 4 of classify] (box1) {};
    \node[basicblock, below = \pad / 4 of parse] (box2) {};

    \draw[myarrow] (OR) -- (classify) node[midway, above, draw=none, font=\footnotesize\sffamily, align=center, xshift=-6pt, yshift=2pt] {voice\\request};
    \draw[myarrow] ([yshift=-5pt]classify.east) -- ([yshift=-5pt]parse.west) node [midway,below,draw=none,font=\footnotesize\sffamily, text width=3em, align=center, yshift=-30pt] {OP type};
    \draw[myarrow] (parse) -- (solve) node [midway,above,draw=none,font=\footnotesize\sffamily, text width=3em, align=center, yshift=15pt] {complete\\OP};
    \draw[myarrow] (assist) -- (solve);
    \draw[myarrow,shorten <=1] ([xshift=-12pt,yshift=-1pt]classify.west) -- ([xshift=-12pt, yshift=27pt]classify.west) -- ([xshift=12pt,yshift=27pt]classify.east) -- ([xshift=12pt,yshift=5pt]classify.east) -- ([yshift=5pt]parse.west) node [midway, above, draw=none, font=\footnotesize\sffamily, text width=5em, align=center, xshift=-4pt, yshift=25pt] {voice\\request};
    \draw[myarrow] (solve) -- (END) node [midway,above,draw=none,font=\footnotesize\sffamily, text width=3em, align=center, xshift=3pt, yshift=8pt] {power\\scheduling vector};
    \draw[myarrow] (sysk) -- ([xshift=-20pt]parse.north);
    \draw[myarrow] (webk) -- ([xshift=20pt]parse.north);

    \begin{scope}[on background layer]
        \node[basicblockshadow,draw=user-request-primary,fill=user-request-gray,inner xsep=8pt,inner ysep=8pt,fit={(classify)($(classify.south)+(0,-10pt)$)},label={[anchor=south west,font=\footnotesize\sffamily\bfseries\color{user-request-primary},outer sep=2pt]south west:Agent 1: OP Classifier}]{};
    \end{scope}

    \begin{scope}[on background layer]
        \node[basicblockshadow,draw=json-primary,fill=json-gray,inner xsep=8pt,inner ysep=8pt,fit={(parse)($(parse.south)+(0,-10pt)$)},label={[anchor=south west,font=\footnotesize\sffamily\bfseries\color{json-primary},outer sep=2pt]south west:Agent 2: Parser}]{};
    \end{scope}

    \begin{scope}[on background layer]
        \node[basicblockshadow,draw=openai-primary,fill=openai-gray,inner xsep=8pt,inner ysep=8pt,fit={(solve)(assist)($(assist.south)+(0,-10pt)$)},label={[anchor=south west,font=\footnotesize\sffamily\bfseries\color{openai-primary},outer sep=2pt]south west:Agent 3: LLM-based OP solver}]{};
    \end{scope}

    \end{tikzpicture}
    \caption{Proposed multi-agent architecture for a voice request to power scheduling vector converter (VRQ2Vec)}
    \label{fig:proposed-architecture}
\end{figure*}

To accommodate the user's demands, the "VRQ2Vec" converter has to initially recognize the user's intent in an accurate way, find the optimum formulation for the corresponding OP, and then solve the latter to generate the recommended power vector that satisfies the requirements. To achieve this goal, we propose a multi-agent architecture \cite{talebirad-MA-2023}, as demonstrated in Fig. \ref{fig:proposed-architecture}. For the problem formulation, we assume a list of most common OPs; for simplicity, we will refer to optimal control problems (OCPs) as OPs as well. The first stage of the VRQ2Vec framework is an \textbf{intent recognition agent (Agent 1)}, which is tasked to identify from the list the most suitable problem that can ideally model the user's intent. The second stage (Parser) is a problem \textbf{parameter identification agent (Agent 2)} which role consists in extracting the parameter information of the selected OP. The third stage corresponds to the \textbf{OP solver agent (Agent 3)}. Note that LLM capabilities are exploited in this study in three ways. First, we exploit their ability to describe a real world problem as a mathematical problem (which is imposed here to be an OP). Second, we exploit the LLM for OP parameter identification purposes. Third, we partially exploit their ability to assist standard OP solvers by allowing the LLM to share their textual knowledge to help better initialize the solver. 
The performance of the three agents and stages of the proposed VRQ2Vec framework will be assessed through two performance metrics. \textbf{Intent Recognition Accuracy (IRA):} assuming the existence of perfect human labeling of every VRQ of a given database into an OP type within a list of OPs, the IRA corresponds to the empirical percentage of OPs properly classified by the LLM-based classifier (Agent 1). \textbf{Average relative optimality loss (AROL)}: knowing that misclassification can occur, the AROL measures how suboptimal is the power vector proposed by the chain Agent 1 $\rightarrow$ Agent 2 $\rightarrow$ Agent 3 in average.



\section{Multi-agent design of the voice request to power vector converter}
\label{sec:architecture}

\subsection{Design of the LLM-based Intent Recognition Agent}
\label{sec:design-classification}

The role of the LLM-based intent recognition agent (\textbf{Agent 1}) is to associate a mathematical problem with a given VRQ from the user. Therefore, it has the role of modeling a speech or text description of a physical problem into equations. Due to the inherent limitations of existing LLMs in describing physical problems as mathematical abstractions, including the latest ChatGPT 4o, we resort to a particular structure of \textbf{Agent 1}, in which this agent is designed to function as a classifier of pre-selected OPs. For simplicity and motivated by the literature of power scheduling problems, we define the following set of possible OPs as described in detail in Table \ref{table:classes-OP}. 
For instance, the VRQ "You have 24h to charge my EV at $80\%$ while minimizing the cost of charging" can be modeled by a linear program (LP) in which: the vector $(c_1,...,c_T)$ (see Table \ref{table:classes-OP}) represents the prices of electricity at time-slots $1,...,T$; choosing $b= -0.8$, $A = (-1, ..., -1)$ translates that the battery state of charge should be at least $80\%$. Note that here we assume 6 OPs but more OPs might be assumed without changing the proposed methodology. However, having more OPs does not necessarily imply having a better performance, showing the importance of both choosing the OPs and the number of OPs. Simulations will support this assertion. The intuition behind the existence of an optimal number of OPs to be used within a given set of OPs for classification is similar to the problem of having less robust digital modulation constellations for large constellations. Here, since the LLM recognition capabilities are not perfect, this introduces "noise" whose impact might be higher when the list of possible OPs gets larger.

Agent 1 therefore uses its language processing abilities to classify the VRQ into an OP (or OCP) type. Note that this classification might be performed by a supervised classical neural network (e.g., an MLP) but this will come at the cost of losing the generalizability capability of LLMs. Rather, we consider an LLM that uses a well-designed context as hard prompting, augmented with function calling abilities. It is designed for the problem of scheduling power for charging an EV but could easily be adapted for the wireless example mentioned in the preceding section. The main LLM we exploit in the performance evaluation part is Llama3  because it is partly open-sourced and because it can be run locally with relatively affordable computational power. To sum up, the design of \textbf{Agent 1} therefore comprises selecting an LLM model (Llama 3 in this paper) and utilizing hard-prompting, with a carefully curated list of OPs (Table \ref{table:classes-OP}) and a well-structured context (Fig.~\ref{fig:prompt}). This setup enables the agent to efficiently achieve the task of classifying the VRQs. To evaluate the performance of \textbf{Agent 1} in terms of IRA, we have constructed a database of VRQs (\textcolor{blue}{\textsf{EVRQ}}); more details will be provided in Sec.~\ref{sec:perf-evaluation}. Note that the LLM's domain knowledge in power scheduling can be further improved through fine-tuning with a specialized dataset, but this is left as an extension of this paper.


\subsection{Design of the LLM-based Parameter Identification Agent}
\label{sec:design-identication}

The classifying agent, \textbf{Agent 1}, identifies a relevant OP type that mathematically describes the request made by the user. In order for \textbf{Agent 3} to be able to numerically solve the corresponding OP, it is necessary first to determine the parameters of the OP. This is where \textbf{Agent 2} plays a role: the OP parameter identifier. One can distinguish between three types of parameters. Type 1: parameters that can be extracted from the VRQ (e.g., the time at which the battery should be recharged to a certain level). Type 2: parameters that are pertinent to the physical system (e.g., the maximum power $x_{\max}$). Type 3: common knowledge parameters that can be sourced from the Internet (e.g., a typical value for the ambient temperature). As far as the design of \textbf{Agent 2} is concerned, the parameters of Types 2 and 3 are considered as inputs to the agent, whereas Type 1 parameters require the LLM-based agent to exploit its language processing skills to extract the relevant parameters from the VRQ. 


To access the parameters determined by the parser, \textbf{Agent 2} requires the capability of function calling, i.e., the ability of passing some parameters to another program by using a particular syntax in order to call a real code function. For each function that needs to be called, we provide a description of the required parameters to the agent.
For the example of EV charging, the parser must behave as follows: initially, it has to extract the time parameters from the user request and then call a function that will initialize those parameters. A time parameter is either an initial time instant, a final time instant, or a duration that can be explicit (8 a.m.) or implicit (tomorrow morning). It is necessary to extract those parameters first, since most of the other parameters are vectors or matrices, whose size depends on the duration of the scheduling. Then, the parser has to call a solving function that depends on the OP type selected, for which the arguments are the parameters of the OP type. In the case of EV charging, to access external parameters, we allow the agent to create the parameters by using attributes and methods from a smart meter. The smart meter is the interface between the environment (or context) and the parser agent. All these parameters form the complete OP that can be passed to the third agent.


\subsection{Design of the LLM-based OP Solving Agent}
\label{sec:OP-solver}

A possible approach to solve the OP which describes the power scheduling problem associated with the VRQ is to use a purely LLM-based agent, that is, to ask an LLM such as GPT 4o or Llama3 to solve the OP. This approach is adopted e.g., by OPRO \cite{OPRO} (OPtimization by PROmpting). Assuming that the optimization task can be described in natural language, OPRO proposes a prompt-based framework to leverage LLMs as numerical optimizers. In each optimization step, the LLM generates new solutions from the prompt that contains previously generated solutions with their values, then the new solutions are evaluated and added to the prompt for the next optimization step. Although promising, these approaches are still limited due to the fact that LLMs were not originally designed to solve mathematical equations\footnote{As per its version of May 31st 2024, Llama 3 8B cannot reliably solve w/o human assistance first-order equations such as $ax + b = 0$ \cite{berman-youtube-2024}.}, and may suffer from accuracy issues, which can be problematic when dealing with stringent physical or quality of service (QoS) constraints. Therefore, instead of trying to design LLM-based optimizers which compete with existing numerical solvers, we rather pursue a \textbf{coupling} approach in which existing solvers are assisted by an LLM. This allows the exploitation of both the determinism/guarantees offered by existing solvers and the creative problem-solving capabilities of LLMs. \textbf{Agent 3} is chosen to be composed of a bank of 6 solvers based on \texttt{scipy.optimize}, \texttt{cvxpy} and \texttt{control} Python libraries: solve\_LP (\texttt{scipy.optimize.linprog}); solve\_QP (\texttt{cvxpy.minimize}); solve\_CP (\texttt{scipy.optimize.minimize}); solve\_MM (\texttt{scipy.optimize.minimize}); solve\_LMT (\texttt{scipy.optimize.milp}); solve\_LQR (\texttt{control.lqr}). The initialization of these solvers are performed by asking an LLM to make the best choice to its knowledge. The values of the parameters which cannot be extracted from the VRQ and the system knowledge are found by using an LLM. 


\section{Performance evaluation}
\label{sec:perf-evaluation}

In this section, we evaluate the performance of the proposed LLM-based approach in EV charging applications. Given the wide diversity of OPs encountered in this domain, we divide them into 6 categories, each with a given type of performance metric that users might consider when initiating a charging session. The main objectives of each category can be listed as follows: Charging cost (CC): Reducing the charging cost; Charging time (CT): Minimizing the time to charge the EV to a target level; Environmental impact (EI): Maximizing the use of renewable energy; Power peak (PP): Minimizing the power peak on the electrical installation; Power variations (PV): Minimizing fluctuations in the power supply to the EV charger; Grid Damage (GD): Limiting the potential damage to the distribution grid installation. Each of these performance metrics is linked to the most suitable OP class that can model the VRQ. To simplify our approach, we assume there is only one OP within each category, as follows:
\begin{table}[ht!]
 \renewcommand{\arraystretch}{1.3}
 \label{tab:corres-op-pm}
 \centering
 \begin{tabular}{|c||c|c|c|c|c|}
 \hline
 Performance metric &CC &CT &PP &PV &GD \\\hline
 OP class &LP &LMT &MM &QP &CP \\\hline
 \end{tabular}
\end{table}

We create knowledge files for each OP class that contains both the description of the problem associated with the performance metrics from the perspective of EV charging, and the generic mathematical description of the OP class independent of EV charging. This comprehensive categorization ensures that we can better instruct LLM agents with prompts for power scheduling in EV charging scenarios.




\subsection{Database and evaluation metric}
To evaluate our framework, we generate a variety of requests that capture multiple types of decisions relevant to different EV charging problems. Specifically, our dataset \textcolor{blue}{\textsf{EVRQ}} includes scenarios such as minimizing charging costs, reducing charging time, maximizing the use of renewable energy, limiting power peaks, reducing power variations, and minimizing grid damage. Additionally, the requests are designed to be either explicit or implicit. Explicit requests are requests that explicitly mention to optimize a particular performance metric, for example: \textit{Charge my EV while minimizing the electricity cost}. However, implicit requests such as \textit{I want my EV to juice up but only when it's financially wise}, specify the performance metric to be assessed with a paraphrase. In total, 800 requests have been generated to compute the IRA, that correspond to 160 requests for each performance metric. To evaluate the performance of our approach, we provide labels for each request, including the ground-truth OP/OCPs class and the optimal solution for the corresponding problem. By comparing these labels with the generated results, the IRA, as defined in Section II, serves as an effective measure of accuracy. In addition, to evaluate the system performance degradation, we consider the average relative optimality loss (AROL), consisting in weighting the optimality loss due to the misclassification by the probability of misclassification, as follows:
\begin{equation}
    \mathrm{AROL}_i=\sum_{J\neq I} { p_{i\rightarrow J} \frac{1}{N} \sum_{n=1}^N {\frac{f_i\left(x_j^{(n)}\right) - f_i\left(x_i^{(n)}\right)}{f_i\left(x_i^{(n)}\right)}}}
\end{equation}
where \(i,j\) are indices for performance metrics with their corresponding OP classes \(I,J\), \(p_{i\rightarrow J}\) represents the probability that a request of type \(i\) is classified as \(J\), $N$ is large and represents the number of random abstract requests to evaluate the average loss for misclassified requests from $i$ to $J$, $x_i^{(n)}$ represents the optimal power vector of the $n$-th request for performance metric \(i\), $x_j^{(n)}$ represents the obtained power vector of the $n$-th request misclassified to $J$ and solved as \(j\), \(f_i\) is the cost function of performance metric \(i\). 
Generating a random abstract request simply consists in randomly selecting a starting time and a duration that will be used to solve the predetermined OPs related to the performance metrics. By taking a large \(N\), we estimate the loss induced by selecting a wrong OP class on average. Then, weighting by the misclassification rates from IRA simulations gives the average relative optimality loss for a particular type of performance metric.


\subsection{Prompt engineering}

\textbf{Basic Prompting:} Use of basic prompt and simple mathematical description of optimization and control problems for classification. In this scenario, we aim to evaluate the performance of the LLM model using a straightforward system prompt. The agent is tasked with classifying the request into an OP class from a predefined list. This setup provides the LLM with only the basic prompts that include the names and mathematical forms of the OP classes, without additional contextual information or guidance related to EV charging.

\textbf{Contextualized Prompting:} improving classification using basic prompts and contextualization of optimization and control problems in the context of EV charging. To assess the impact of augmenting the LLM with more detailed knowledge, we extend the basic system prompt from Scenario 1 by appending comprehensive knowledge files. These files provide both textual and mathematical descriptions of typical EV charging problems within each OP class. By enriching the LLM's input with this detailed information, we aim to improve its ability to recognize the appropriate OP class and accurately translate the EV charging request into a canonical form suitable for external solvers.

\textbf{Error-Informed Prompting:} improving classification by analyzing the previous errors. In this scenario, we further refine the system prompt to enhance the classifier's performance. This involves incorporating specific remarks to guide the LLM more effectively. These remarks are mainly obtained from analyzing the mistakes made with the two prompting techniques mentioned above, as well as leveraging expertise in EV charging and optimization to identify descriptions with textual ambiguity and sources of optimization confusion. By regulating the structure and content of the prompts, we aim to streamline the LLM's decision-making process, ensuring more accurate and reliable classification and problem formulation.

These scenarios illustrate the progressive enhancements in prompt engineering techniques, demonstrating how additional knowledge and refined guidance can significantly improve the performance of LLMs in classifying and solving complex OPs in EV charging applications, as shown in the following figures.



\begin{figure}
    \centering
    \includegraphics[width=\linewidth]{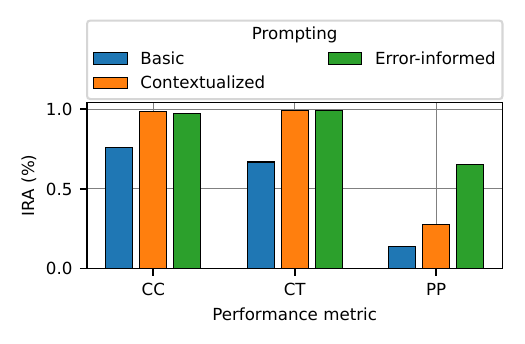}
    \caption{Influence of context knowledge for Agent 1 in terms of IRA.
    The evaluation is performed for 3 types of voice requests (CC, CT, and PP). The gains provided by knowledge files are seen to be significant.}
    \label{fig:ira-comp-scenarios}
\end{figure}

\begin{figure}
    \centering
       \includegraphics[width=\linewidth]{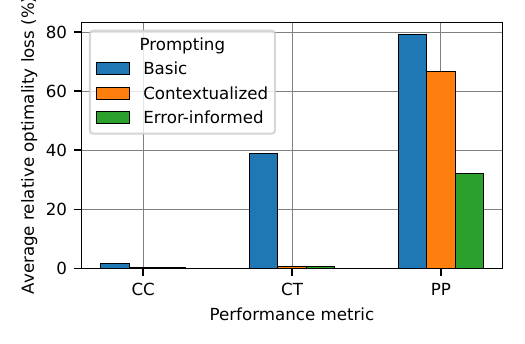}
    \caption{Influence of context knowledge in terms of optimality loss for the final power scheduling performance metric.}
    \label{fig:opt-loss-comp-scenarios}
\end{figure}

\begin{figure*}
     \centering
      \includegraphics[width=\linewidth]{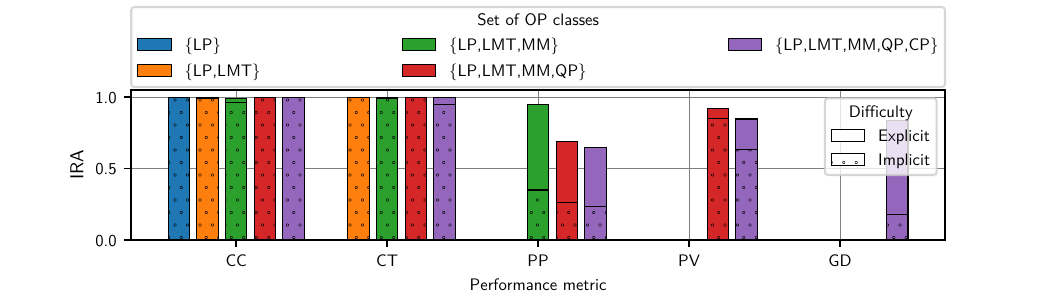}
     \caption{For 5 different performance metrics (CC,...,GD): influence of the set of selected OPs ($\{\mathrm{LP}, \mathrm{LMT}\}$ means for instance that either Linear Programming or Linear Minimum Time has to be chosen) on its ability to select the most suitable OP class.}
     \label{fig:ira-evo-op}
\end{figure*}

\newcolumntype{Y}{>{\centering\arraybackslash}X} 
\setlength{\tabcolsep}{6pt} 

\begin{table*}[ht!]
    \renewcommand{\arraystretch}{0.8} 
    \caption{Description of Optimization (and optimal control) Problem classes considered by Agent 1}
    \label{table:classes-OP}
    \centering
    \normalsize 
    \begin{tabularx}{\textwidth}{|Y|Y|Y|}
        \hline
        \textbf{Linear Programming (LP)} & \textbf{Quadratic Programming (QP)} & \textbf{Mini-Max Class (MM)} \\
        \hline
        \vspace{1pt}
        \(
        \begin{aligned}
            & \underset{x}{\text{minimize}} &\quad &c^\top x \\
            & \text{s.t.} &\quad &A x \leq b \\
            &             &\quad &A_\text{eq} x = b_\text{eq} \\
            &             &\quad &{0} \leq x_{\text{min}} \leq x_t \leq x_{\text{max}}
        \end{aligned}
        \)
        \vspace{1pt}
        &
        \vspace{1pt}
        \(
        \begin{aligned}
            & \underset{x}{\text{minimize}} &\quad &\frac{1}{2} x^\top Q x + c^\top x \\
            & \text{s.t.} &\quad &A x \leq b \\
            &             &\quad &A_\text{eq} x = b_\text{eq} \\
            &             &\quad &{0} \leq x_{\text{min}} \leq x_t \leq x_{\text{max}}
        \end{aligned}
        \)
        \vspace{1pt}
        &
        \vspace{1pt}
        \(
        \begin{aligned}
            & \underset{x}{\text{minimize}} &\quad &\max_i f_i(x) \\
            & \text{s.t.} &\quad &A x \leq b \\
            &             &\quad &A_\text{eq} x = b_\text{eq} \\
            &             &\quad &{0} \leq x_{\text{min}} \leq x_t \leq x_{\text{max}}
        \end{aligned}
        \)
        \vspace{1pt}
        \\
        \hline
        \textbf{Convex Programming (CP)} & \textbf{Linear Minimum-Time (LMT)} & \textbf{Linear Quadratic Regulator (LQR)} \\
        \hline
        \vspace{1pt}
        \(
        \begin{aligned}
            & \underset{x}{\text{minimize}} &\quad &f(x) \\
            & \text{s.t.} &\quad &g_i(x) \leq 0, \quad i=1\ldots m \\
            &             &\quad &A_\text{eq} x = b_\text{eq} \\
            &             &\quad &{0} \leq x_{\text{min}} \leq x_t \leq x_{\text{max}}
        \end{aligned}
        \)
        \vspace{1pt}
        &
        \vspace{1pt}
        \(
        \begin{aligned}
            & \underset{x}{\text{minimize}} &\quad &\tau \\
            & \text{s.t.} &\quad &s_{t+1} = A s_t + B x_t \\
            &             &\quad &s_0 = s_i, s_{\tau} = s_f \\
            &             &\quad &{0} \leq x_{\text{min}} \leq x_t \leq x_{\text{max}} \\
            &             &\quad &s_{\text{min}} \leq s_t \leq s_{\text{max}}
        \end{aligned}
        \)
        \vspace{1pt}
        &
        \vspace{1pt}
        \(
        \begin{aligned}
            & \underset{x}{\text{minimize}} &\quad &\sum_{t=0}^{N-1} (s_t^\top Q s_t + r x_t^2) \\
            &                               &\quad &+ s_N^\top Q_f s_N \\
            & \text{given $s_0$ s.t.} &\quad &{0} \leq x_{\text{min}} \leq x_t \leq x_{\text{max}} \\
            &             &\quad &s_{t+1} = A s_t + B x_t\\
        \end{aligned}
        \)
        \vspace{1pt}
        \\
        \hline
    \end{tabularx}
    \vspace{10pt}
    \begin{tabularx}{\textwidth}{|X|}
        \hline  \textbf{Notations:} $x$: power scheduling vector; $x_t, s_t$ : power and state at time $t$; other quantities are parameters. 
        \\
        \hline
    \end{tabularx}
\end{table*}
\subsection{Simulation results}

We implemented the different agents in Python by using the \emph{ollama} library and Llama3 8B as the base model. We set the temperature of the model to 0 to eliminate any randomness, ensuring that each request is consistently classified into the same OP class. Due to the computational constraints of using Llama3 8B, we limited the scope of some simulations by reducing the number of OP classes known to Agent 1 to three.

Fig.~\ref{fig:ira-comp-scenarios} illustrates the IRA performance with respect to the different proposed prompting techniques. The obtained results show a clear trend where error-informed prompting achieves the highest IRA across all performance metrics. Meanwhile, by providing the typical EV charging problems specific to each OP class, contextualized prompting shows a moderate improvement over basic prompting, where the latter is application agnostic. To further study the impact of different prompting schemes, in Fig.~\ref{fig:opt-loss-comp-scenarios} we demonstrate the average optimality loss using the three different prompting techniques. Confirming the earlier results, error-informed prompting exhibits the lowest average optimality loss, indicating the most accurate problem formulation. These two figures corroborate the advantages of utilizing advanced  prompting techniques in the underlying framework to notably enhance the model's ability to classify requests accurately, as well as to introduce improved accuracy performance with respect to the final charging power vector.

Fig.~\ref{fig:ira-evo-op} illustrates the IRA performance depending on the number of OP  classes provided to the classifier. With only one OP class (LP), the classifier can only handle CC requests, as other requests cannot be resolved using linear programming alone. This explains the absence of the blue bar in the chart for requests outside the CT category, as the IRA is zero when relying only on LP. In addition, the figure highlights that for certain requests, such as those in the PP category, the IRA decreases as the number of OP classes increases. Thus, while a limited number of OP classes restricts the number of treatable requests, adding more OP classes negatively impacts the IRA. It also emphasizes the importance of explicit user requests, as seen with PP requests where providing 3 OPs results in a significant gap in IRA (90\% compared to 35\% accuracy), between implicit and explicit requests. A similar pattern is observed for GD requests. This indicates that detailed and explicit user requests are essential to achieve higher accuracy in classification. It can be further observed from the figure that classifiers with different sets of OP classes perform very well in CC and CT categories, compared to other ones. This is motivated by the fact that cost and time are common requests by the users and are relatively easier to be classified, for both explicit and implicit requests.

\begin{figure}
    \centering
    \includegraphics[angle=0,width=\linewidth]{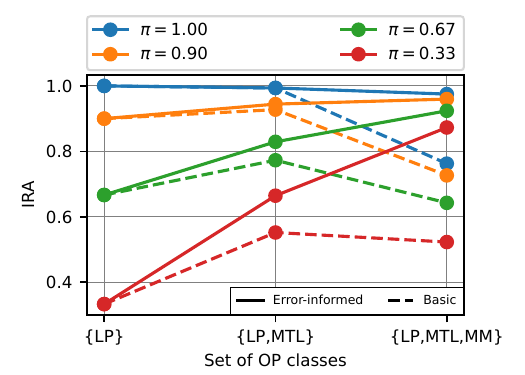}
    \caption{Impact of the set size of OPs on the classification accuracy of Agent 1. The result is seen to depend on the distribution of the voice request database (which is represented by the probability $\pi$ of having a request of type CC). Having a larger set of OPs to model the power scheduling problems can negatively affect the recognition accuracy.}
    \label{fig:cardinality}
\end{figure}

To further understand other metrics that impact OP classification accuracy, Fig.~\ref{fig:cardinality} explores the influence of a set of selected OPs on Agent 1. The IRA is plotted versus different cardinality levels  with varying distributions of request categories. The probabilities are defined such that a request belonging to CC, CT, and PP is defined as $\pi, (1-\pi)/2,(1-\pi)/2$, respectively. When requests are uniformly distributed across all categories, using a larger cardinality improves the IRA significantly. This demonstrates that well-designed prompts can handle a broader range of classes effectively, hence, enhancing classification accuracy. When requests are predominantly from one category (with $\pi$ close to 1), a smaller cardinality can yield better IRA. Additionally, in scenarios where requests are evenly spread, advanced prompts can help the LLM to distinguish between different OPs efficiently, whereas basic prompts are sufficient for an acceptable classification accuracy when requests are very concentrated.

To assess the effect of the selected LLM model in this framework, we tested the error-informed prompting with other LLM models, including GPT-4o, Gemini Advanced (GA), and Llama 3 70B, and examined the classification accuracy on a sample of 30 requests from our database. These samples comprise three distinct sets: 10 requests that were correctly classified by Llama3 8B to verify if other models can perform at least as well as Llama3 8B, 10 requests that were misclassified by Llama3 8B to evaluate if the other models can improve upon Llama3's performance, and 10 EI requests that are not included in the classifier's knowledge base to examine if the selected models can extrapolate and classify requests that are not directly within their known dataset. The obtained IRA results for different models are outlined in Table.~\ref{table:different-llms}. The results demonstrate the potential of sophisticated/larger LLMs in improving the OP classification. It further shows that the proposed framework is well-suited for most popular LLMs, emphasizing on the generalizability of this proposed architecture. Also, it suggests that such performance can be further enhanced with better-trained models.

\begin{figure}
    \centering
\includegraphics[angle=0,width=\linewidth]{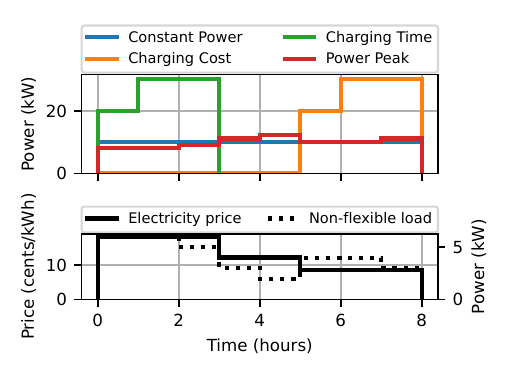}
    \caption{Final power scheduling vector generated by the proposed agent chain for different classes of voice requests (CC, CT, PP) compared to a basic constant power charging policy. The results constitute a \textbf{proof of concept} for the proposed methodology when applied to EV charging: user's voice requests are translated to a very suitable power scheduling policy for very diverse requests (800 VRQs).}
    \label{fig:profiles}
\end{figure}

Finally, in Fig.~\ref{fig:profiles}, we illustrate the patterns of the yielded power vectors for different requests types. For charging cost minimization requests, the charging power is primarily allocated to time slots with lower electricity prices. In contrast, for requests aiming to minimize the charging time, the power vector is more concentrated at the beginning, regardless of other metrics. For power peak minimization requests, the charging power is higher during time slots with lower non-flexible loads to balance the overall electrical load. By adapting to the charging requests, the power vector can be adjusted to meet very diverse requirements effectively.


\begin{table}[ht!]
\renewcommand{\arraystretch}{1.3} 
\caption{Influence of the LLM model on accuracy (IRA).} 
\label{table:different-llms}
\centering
\begin{tabular}{|c||c|c|c|c|}
\hline
& \textbf{Llama3 8B} & \textbf{GPT-4o} & \textbf{GA} & \textbf{Llama3 70B} \\
\hline
\textbf{Perfectly} & 100\% & 100\% & 100\% & 100\% \\
\textbf{Classified} & & & & \\
\hline
\textbf{Misclassified} & 0\% & 90\% & 60\% & 90\% \\
\hline
\textbf{EI} & 90\% & 100\% & 100\% & 90\% \\
\hline
\end{tabular}
\end{table}


\section{Conclusion}
\label{sec:conclusion}
This paper proposes, for the first time in the literature, how to exploit LLMs to convert an arbitrary VRQ into a power vector. We develop an efficient multi-agent architecture that relies on existing LLM models. We corroborate the efficacy of the proposed methodology by a thorough performance analysis for the EV charging problem. To conduct this analysis, we create a database of VRQs and proposed approaches to handle practical implementation aspects, including OP parameter identification. The corresponding results provide key insights. Having a larger set of possible OPs to model a real-world problem can be detrimental to the model choice problem, creating a tradeoff between accurately modeling a physical problem and the model's ability to correctly recognize the type of problem. The proposed architecture is original and opens a broad avenue for improvements. In particular: the design of the agent which models the physical problem can be improved by having a more diverse set of OPs and by being fine-tuned for wireless/energy networks; the agent which solves the selected OP can be improved by better coupling the capabilities of standard OP solvers with creative LLM-based solvers. We believe that the approach introduced in this paper will be key for humans to interact with wireless/energy networks. 


\section{Appendix}
\label{sec:appendix}

\begin{figure}[h!]
    \centering
    \resizebox{\linewidth}{0.73\linewidth}{\InstructBoxFile{Files/instructions/classifier-sp2.txt}}
    \caption{Extract from a system prompt given to Agent 1.}
    \label{fig:prompt}
\end{figure}

\bibliographystyle{IEEEtran}

\end{document}